\documentclass[final]{cvpr}

\usepackage{times}
\usepackage{epsfig}
\usepackage{graphicx}
\usepackage{amsmath}
\usepackage{amssymb}
\usepackage{bm}
\usepackage{multirow}
\usepackage{soul}
\usepackage{subcaption}
\usepackage{stfloats}
\usepackage{float}

\newlength\savewidth\newcommand\shline{\noalign{\global\savewidth\arrayrulewidth
		\global\arrayrulewidth 1pt}\hline\noalign{\global\arrayrulewidth\savewidth}}

\usepackage[pagebackref=true,breaklinks=true,colorlinks,bookmarks=false]{hyperref}

\def\confYear{CVPR 2021}

\begin{document}

\title{Learning Transferable Pedestrian Representation from  Multimodal\\ Information Supervision}

\author{Liping Bao$^1$\quad
		Longhui Wei$^1$\quad
		Xiaoyu Qiu$^1$\quad
		Wengang Zhou$^1$\quad \\
		Houqiang Li$^1$\quad 
		Qi Tian$^2$
		\and
		$^1$University of Science and Technology of China \quad $^2$Huawei Cloud  \\
		{\tt\small \{baoliping, qiuxy\}@mail.ustc.edu.cn} \quad {\tt\small weilh2568@gmail.com} \\
		{\tt\small \{zhwg, lihq\}@ustc.edu.cn} \quad {\tt\small tian.qi1@huawei.com}
	}
\maketitle

\begin{abstract}
Recent researches on unsupervised person re-identification~(reID) have demonstrated that pre-training on unlabeled person images achieves superior performance on downstream reID tasks than pre-training on ImageNet.
However, those pre-trained methods are specifically designed for reID and suffer flexible adaption to other pedestrian analysis tasks.
In this paper, we propose VAL-PAT, a novel framework that learns transferable representations to enhance various pedestrian analysis tasks with multimodal information. 
To train our framework, we introduce three learning objectives, \emph{i.e.,} self-supervised contrastive learning, image-text contrastive learning and multi-attribute classification.
The self-supervised contrastive learning facilitates the learning of the intrinsic pedestrian properties, 
while the image-text contrastive learning guides the model to focus on the appearance information of pedestrians. 
Meanwhile, multi-attribute classification encourages the model to recognize attributes to excavate fine-grained pedestrian information. 
We first perform pre-training on LUPerson-TA dataset, where each image contains text and attribute annotations, and then transfer the learned representations to various downstream tasks, including person reID, person attribute recognition and text-based person search.
Extensive experiments demonstrate that our framework facilitates the learning of general pedestrian representations and thus leads to promising results on various pedestrian analysis tasks. 
\end{abstract}
\vspace{-4mm}

\begin{figure}[t]
  \centering
  \includegraphics[width=\linewidth]{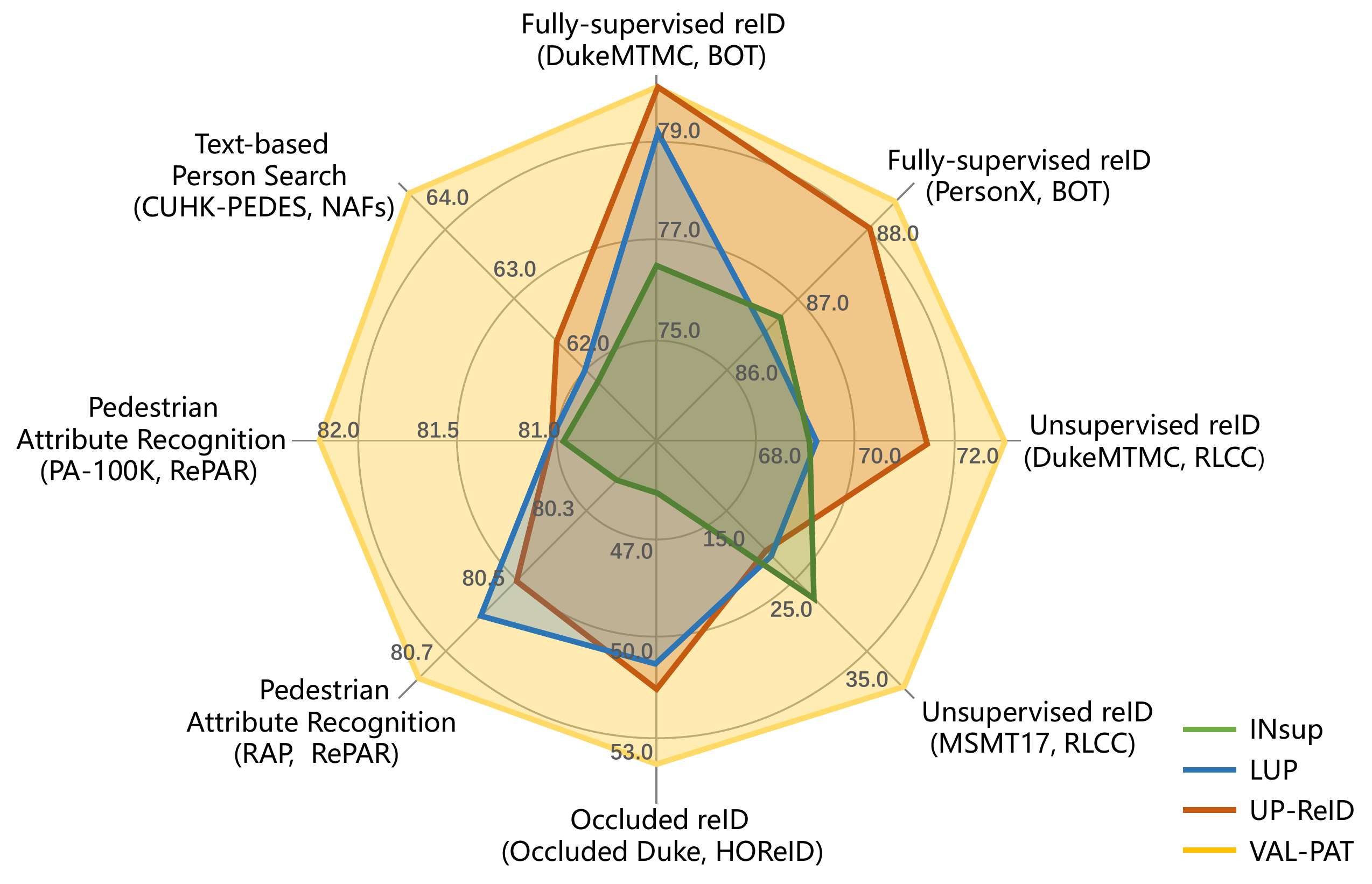}
   \caption{VAL-PAT achieves the best performance on a series of downstream pedestrian analysis tasks compared with other pre-trained models.
   ``INsup" refers to the supervised pre-trained model on ImageNet.
   ``LUP'' refers to MoCo v2 pre-trained on LUPerson.
   ``UP-ReID'' represents the existing best pre-trained model designed for person reID.
}
   \label{fig1} 
   \vspace{-3mm}
\end{figure}
\section{Introduction}
\label{sec:intro}
As an emerging research topic in computer vision, pedestrian analysis has attracted substantial attention thanks to its significant potential in video surveillance applications.
Generally, pedestrian analysis involves many tasks, 
such as person re-identification~(reID)~\cite{zhao2017spindle, zhao2017deeply, sun2018beyond, yao2019deep}, pedestrian attribute recognition~\cite{tang2019improving, jia2021rethinking, jia2021spatial}, and text-based person search~\cite{li2017person, sarafianos2019adversarial, shao2022learning, suo2022simple}.
With the rise of deep learning, the state-of-the-art methods on pedestrian analysis heavily rely on deep neural networks trained with large amounts of annotated data~\cite{wei2018person,li2016richly,liu2017hydraplus,li2017person},
which is a non-trivial issue, 
as it involves annotating fine-grained pedestrian attributes or associating person images captured by different cameras.
With limited labeled data, most of current methods~\cite{sun2018beyond,he2021transreid,yao2019deep,dai2019batch} are firstly pre-trained on ImageNet and then fine-tuned on downstream pedestrian datasets.
Nonetheless, those solutions suffer a domain gap between the ImageNet~\cite{deng2009imagenet} and the pedestrian datasets, since the former consists of general images and focuses on category-level classification while the latter only contains pedestrain images and focuses on instance-level classification.

Recently, self-supervised learning~\cite{he2020momentum,chen2020simple,he2022masked,bao2021beit} has shown remarkable success in visual pre-training tasks.
Inspired by these advances, Fu et al.~\cite{fu2021unsupervised} make the pioneering attempt at large-scale unsupervised pre-training for person reID and establish a large-scale unlabeled person reID dataset, LUPerson.
However, this method directly utilizes MoCo v2~\cite{chen2020improved}, a self-supervised method designed for the ImageNet, without taking into account the unique characteristics of the person reID tasks.
To address this issue, UP-ReID~\cite{yang2022unleashing} is further developed to consider the global image-level and local patch-level features simultaneously, allowing for the extraction of fine-grained clues from person images.
It is worth noting that these self-supervised methods are designed specifically for person reID and may suffer flexible adaption to other pedestrian analysis tasks such as person attribute recognition and text-based person search.
As shown in Figure~\ref{fig1}, LUP and UP-ReID only achieve marginal performance improvements over INsup on person attribute recognition and text-based person search tasks.

To learn general pedestrian representations that transfer well to various pedestrian analysis tasks, we introduce a novel multimodal learning framework trained with Vision, Attribute and Language supervisions, named as VAL-PAT.
Unfortunately, none of the existing large-scale pedestrian datasets contain both text and attribute annotations.
Therefore, we construct a multimodal pedestrian dataset LUPerosn-TA,
where each image is paired with a corresponding textual description and attribute annotations.
Instead of crawling pedestrian image-text pairs from Web, we utilize a model on image captioning~\cite{xu2015show} to generate textual descriptions for the images in LUPerson,
and then extract common nouns and adjectives from each textual description as attributes for its corresponding image.

Based on the aforementioned multimodal pedestrian dataset, we introduce three optimization objectives to train our proposed framework.
We first employ self-supervised contrastive loss~(SSL) to minimize the distance between different views of the same image while maximizing the distance between different images, with the aim of extracting the inherent properties of the pedestrian images.
We also use image-text contrastive loss~(ITC) to pull matched image-text pairs closer while pushing away non-matched ones.
Since texts provide comprehensive and detailed descriptions of the pedestrians' appearance, ITC encourages the model to prioritize information about the pedestrians' appearance, while disregarding unnecessary information like background.
Additionally, a multi-attribute classification loss~(MAC) is introduced to guide the model to recognize the attributes presented in the image-text pairs, thereby facilitating the extraction of fine-grained information.
Incorporating these three objectives into a single multimodal learning framework can enable the candidate pre-trained model to learn more general and robust representations that can easily transfer to various pedestrian-related tasks.


In summary, our major contributions are three-fold as follows:
\vspace{-2mm}
\begin{itemize}
    \item We introduce a simple but effective framework, VAL-PAT, optimized with three objectives, \emph{i.e.,} image self-supervised contrastive loss, image-text contrastive loss and  multi-attribute classification loss.
    To our best knowledge, this is the first attempt to learn general representations 
    that transfer well to various pedestrian analysis tasks.
    \vspace{-2mm}
    \item To train the framework, we construct a large-scale multimodal pedestrian dataset, LUPerson-TA, 
    where each image corresponds to a textual description and attribute annotations.
    \vspace{-2mm}
    \item Extensive experiments on various pedestrian analysis tasks 
    reveal that our VAL-PAT achieves superior transfer performances on most of pedestrian datasets, which validates the effectiveness of our proposed framework.
\end{itemize}

\section{Related Works}
\subsection{Visual Representation Learning}
In the past two decades, supervised learning~\cite{NIPS2012_c399862d,he2016deep} has played a dominant role in visual representation learning.
Pre-trained weights on ImageNet have been widely used in various downstream vision tasks, such as object detection~\cite{ren2015faster,redmon2016you,liu2016ssd}, semantic segmentation~\cite{he2017mask,long2015fully,chen2017deeplab}, person reID~\cite{sun2018beyond,he2021transreid,dai2019batch,zhang2020relation}, \emph{etc.}, to alleviate the problem of insufficient labeled data. 
Recently, the prevailing trend has been disrupted by self-supervised learning based on contrastive learning~\cite{he2020momentum,chen2020simple,wang2021dense} and masked image modeling~\cite{bao2021beit,he2022masked,xie2022simmim}, 
which learns better transferable representations with only unlabeled images.
Contrastive learning takes the discriminative learning paradigm by maximizing the similarity between different views of the same image while minimizing the similarity between different images. 
Differently, masked image modeling works in a generative way by reconstructing the masked part of the original input from the given visible parts.

Instead of image self-supervision, some researchers have explored language as supervision to learn meaningful visual representations.
For example, VirTex~\cite{desai2021virtex} has demonstrated that natural language can provide more data-efficient supervision for learning transferable visual representations.
Recently, CLIP~\cite{radford2021learning} and ALIGN~\cite{jia2021scaling} have achieved impressive zero-shot classification performance by leveraging image-text contrastive learning on 400M/1B image-text pairs.
Furthermore, combining image self-supervision and language supervision, SLIP~\cite{mu2022slip} significantly improves performances in the zero-shot transfer, linear classification and fine-tuning settings compared to CLIP and SIMCLR~\cite{chen2020simple}, showing language can effectively aid image self-supervised learning to learn better representations.


\begin{figure*}[t]
  \centering
  \includegraphics[width=\linewidth]{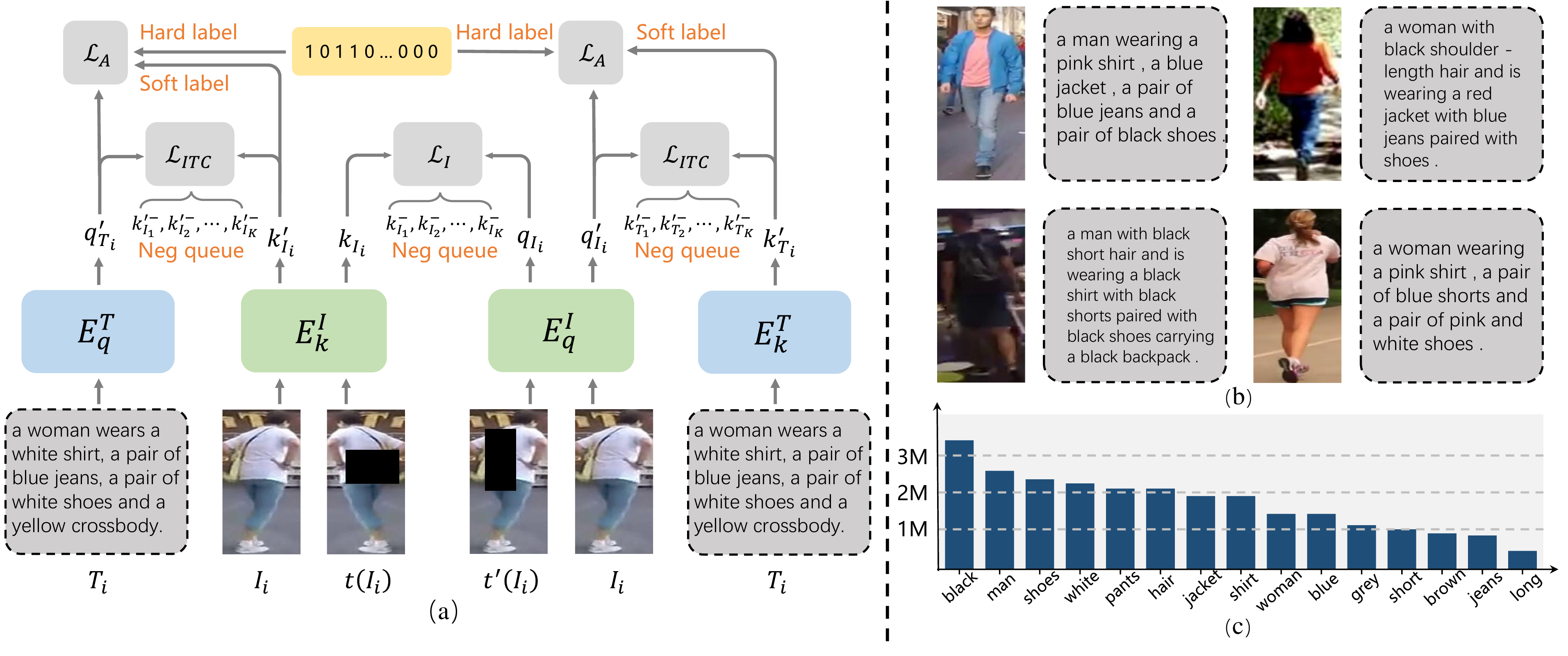}
  \vspace{-7mm}
   \caption{Illustration of our pre-training framework and LUPerson-TA dataset.
    (a) is the overview of our framework, optimized with image self-supervised contrastive loss $\mathcal{L}_{I} $, image-text contrastive loss $\mathcal{L}_{ITC} $ and multi-attribute classification loss $\mathcal{L}_{A} $.
   We leverage the logits computed over $ k^\prime_{T_{i}} $ and $ k^\prime_{I_{i}} $ as soft labels to alleviate the impact of noise attributes while computing $\mathcal{L}_{A} $.
   (b) shows several image-text pairs in the LUPerson-TA, where the sentences can describe pedestrians' appearance in details.
   (c) is a histogram of the top-15 attributes. 
   The ``M'' represents million.
  }
   \label{fig2} 
  \vspace{-4mm}
\end{figure*}

\subsection{Unsupervised Pretraining for Person Re-identification}
\vspace{-2mm}
Limited by the size of labeled reID datasets, previous works~\cite{dai2019batch,sun2018beyond,he2021transreid} usually load weights pre-trained on ImageNet before training on reID datasets.
However, it may be unreasonable that there exists a large domain gap between ImageNet and reID datasets.
Recently, Fu et al.~\cite{fu2021unsupervised} first establish a large-scale unlabeled person reID dataset, LUperson, and verify that pre-training on LUPerson significantly improves performances on downstream reID tasks compared to pre-training on ImageNet.
Luo et al.~\cite{luo2021self} extensively investigate the influence of different model architectures and pre-trained methods and demonstrate that the transformer-based DINO~\cite{caron2021emerging} can achieve better performance.
However, these works directly utilize self-supervised learning methods designed for ImageNet classification without specialized adaption for reID tasks.
Yang et al.~\cite{yang2022unleashing} propose a reID-specific pre-training framework UP-ReID and introduce global consistency and intrinsic contrastive constraints to learn augmentation-invariant and fine-grained representations.
Concurrently, Zhu et al.~\cite{zhu2022pass} propose PASS based on DINO, which automatically extracts part-level features to offer fine-grained information.
Nevertheless, these pre-training models are tailored for reID tasks and may not be transferred to other pedestrian-related tasks, such as pedestrian attribute recognition, text-based person search, \emph{etc}.
In this paper, we propose a novel multimodal learning framework that utilizes multimodal supervision to guide model to learn general pedestrian representations beneficial for a series of downstream pedestrian analysis tasks.

\section{VAL-PAT Framework}
In this section, we present our proposed multimodal learning framework VAL-PAT in details.
As shown in Figure~\ref{fig2}(a), VAL-PAT is optimized using image self-supervised contrastive loss, image-text contrastive loss and multi-attribute classification loss.
To this end, text and attribute annotations are required for each pedestrian image.
Unfortunately, none of the existing pedestrian datasets fully satisfy these requirements.
Therefore, we first introduce how to construct a multimodal pedestrian dataset, where each image is paired with a corresponding textual description and attribute annotations.
Then, we elaborate on each optimization objective and present the full pre-training objective.

\subsection{LUPeron-TA Dataset}
The typical approaches\cite{radford2021learning,jia2021scaling,schuhmann2021laion} to collect numerous image-text pairs are through crawling from Web.
However, crawling image-text pairs featuring pedestrian is more challenging due of the limitation of image category and the requirements for high-quality text that can describe the pedestrians' appearance in details.
To tackle this issue, we develop a simple method based on LUPerson dataset to construct a pedestrian dataset with text and attribute annotations and name this dataset as LUPerson-TA.
We first train an image captioning model~\cite{xu2015show} from scratch on the CUHK-PEDES~\cite{li2017person} and ICFG-PEDES~\cite{ding2021semantically} datasets, which are commonly used in text-based person search tasks and where each image has one or two textual descriptions.
Next, we utilize the trained model to generate textual descriptions for each image in the LUPerson dataset.
We provide some examples of image-text pairs in Figure~\ref{fig2}(b),
where these texts describe the characteristics of pedestrians in details.
To obtain attribute annotations, we extract all nouns and adjectives from these descriptions using the NLTK tagger~\cite{bird2009nltk} and retain only the most frequently occurring ones, resulting in a list of 1359 attributes. 
The histogram of the top-15 attributes is demonstrated in Figure~\ref{fig2}(c).

\subsection{Self-supervised Contrastive Learning}
One of the representative paradigms in visual self-supervised learning is contrastive learning~\cite{he2020momentum,chen2020simple,wu2018unsupervised}, which can be considered as an instance classification task where each image is treated as a separate class. 
Instead of traditional classifier, contrastive learning leverages the ``noise contrastive estimator~(NCE)~\cite{gutmann2010noise}'' to maximize the similarity between different augmented views of the same image and minimize the similarity between different images.
Specifically, given an image $ I_{i} $ and two separate data augmentation operators $ t $ and $ t^\prime $ sampled from the same family of augmentations, we can obtain two augmented views $ t\left(I_{i}\right) $ and $ t^\prime\left( I_{i}\right) $ of the image $ I_{i} $, one encoded by image encoder $ E_{q}^{I} $ as $ q_{I_{i}} $ and the other encoded by image encoder $ E_{k}^{I} $ as $ k_{I_{i}} $, where $ E_{k}^{I} $ is the momentum updating version of $ E_{q}^{I} $.
Moreover, we maintain a queue of negative samples $ \left\{ k_{I_{1}}^{-}, k_{I_{2}}^{-}, \cdots, k_{I_{K}}^{-}\right\} $, which are encoded by $ E_{k}^{I} $ over augmented views from different images.
Formally, the image self-supervised contrastive loss is defined as: \par
\vspace{-2mm}
\begin{small}
\begin{equation}
\label{eqn:contrastive}
    \mathcal{L}_{I} = - \text{log}\frac{\text{exp}(q_{I_{i}} \cdot k_{I_{i}} / \tau)}{\text{exp}(q_{I_{i}}\cdot k_{I_{i}}/\tau) + \sum_{j=1}^{K} \text{exp}(q_{I_{i}}\cdot k_{I_{j}}^{-}/\tau)},
\end{equation}
\end{small}

\vspace{-1mm}
\noindent
where $ \tau $ is a temperature hyper-parameter, and $K$ is the number of negative samples.

\subsection{Image-text Contrastive Learning}
CLIP~\cite{radford2021learning} and ALIGN~\cite{jia2021scaling} have demonstrated that utilizing contrastive loss to maximize the similarity between the matched image-text pairs, while minimizing the similarity between the non-matched pairs, encourages the model to learn rich semantic visual representations.
To be specific, given an image-text pair $\{I_i, T_i\}$, we adopt two modality-specific encoders $E_{q}^{I}$ and $E_{q}^{T}$ to extract the image query $q^\prime_{I_{i}}$ and text query $q^\prime_{T_{i}}$, respectively. 
Simultaneously, image key $k^\prime_{I_{i}}$ and text key $k^\prime_{T_{i}}$ are obtained by using $E_{k}^{I}$ and $E_{k}^{T}$, which are momentum updating versions of $E_{q}^{I}$ and $E_{q}^{T}$, respectively. 
Similar to images self-supervised contrastive loss, we also maintain two negative queues, the image negative queue $\left\{k_{I_{1}}^{\prime-}, k_{I_{2}}^{\prime-}, \cdots, k_{I_{K}}^{\prime-}\right\}$ and the text negative queue $\left\{k_{T_{1}}^{\prime-}, k_{T_{2}}^{\prime-}, \cdots, k_{T_{K}}^{\prime-}\right\}$, to expand the number of negative samples.
The contrastive loss, in the scenario of image-text pairs, is given as:  \par
\begin{small}
\begin{align}
	\mathcal{L}_{I2T} = - \text{log}\frac{\text{exp}(q^\prime_{I_{i}} \cdot k^\prime_{T_{i}} / \tau^\prime)}{\text{exp}(q^\prime_{I_{i}}\cdot k^\prime_{T_{i}}/\tau^\prime) + \sum_{j=1}^{K} \text{exp}(q^\prime_{I_{i}}\cdot k^{\prime-}_{T_{j}}/\tau^\prime)},\nonumber\\
	\mathcal{L}_{T2I} = - \text{log}\frac{\text{exp}(q^\prime_{T_{i}} \cdot k^\prime_{I_{i}} / \tau^\prime)}{\text{exp}(q^\prime_{T_{i}}\cdot k^\prime_{I_{i}}/\tau^\prime) + \sum_{j=1}^{K} \text{exp}(q^\prime_{T_{i}}\cdot k^{\prime-}_{I_{j}}/\tau^\prime)},
  \label{equ:itc_loss}
\end{align}
\end{small}

\vspace{-2mm}
\noindent
where $\tau^\prime$ is the temperature to scale the logits, which is learned together with all other parameters.

\subsection{Multi-Attribute Classification}
While self-supervised contrastive learning and image-text contrastive learning can result in powerful representations, they focus solely on exploring the global features of images and texts, potentially neglecting significant fine-grained information.
Motivated by this, we introduce a multi-attribute classification objective, which is a multi-label classification task. 
By training model to recognize attributes (\emph{e.g.}, backpack, tie, watch) in both images and texts, the model can mine fine-grained local clues and learn more discriminative representations.
Moreover, as illustrated in Figure~\ref{fig2}(a), applying multi-attribute classification to image and text encoders encourages both encoders to capture the same fine-grained attributes, facilitating fine-grained alignment between image and text modalities.

Formally, given an attribute label $ \bm{y}_i \in \left\{0,1\right\}^M$ and an attribute classifier $ W_q $ with bias $ \bm{z}_q $, where $ M $ denotes the number of attributes and the zeros and ones indicate the absence and presence of the corresponding attribute, we adopt the widely used binary cross-entropy loss in pedestrian attribute recognition~\cite{jia2021spatial} as our optimization target: \par
\vspace{-4mm}
\begin{small} 
\begin{equation}
\begin{split}
    \mathcal{L}_{A}^H = \sum_{j=1}^N \Big( & w_j\left( y_{i, j}\log p_{i, j}^I + \left(1 - y_{i, j}\right) \log \left(1 - p_{i, j}^I\right)\right) \\
                           + & w_j\left( y_{i, j}\log p_{i, j}^T + \left(1 - y_{i, j}\right) \log \left(1 - p_{i, j}^T\right)\right) \Big), \nonumber
  \label{equ:ac_loss}
\end{split}
\end{equation}
\end{small}
\begin{equation}
    p_{i}^I = \sigma\left(W_q q^\prime_{I_i} + \bm{z}_q \right), p_{i}^T = \sigma\left(W_q q^\prime_{T_i} + \bm{z}_q \right),
  \label{equ:ac_loss1}
  \vspace{1mm}
\end{equation}
where $ \bm{p}_{i,j}^I $ and $ \bm{p}_{i,j}^T $ is the prediction probability of $ j $-th attribute, $ \sigma\left(\cdot\right) $ is the sigmoid function and $ w_j $ is the attribute weight of $ j $-th attribute to alleviate the distribution imbalance between attributes.

However, since the mined attribute labels are noisy, using the hard binary attribute labels $ y_i $ as supervision causes the model to overfit to noise, which significantly harms to transferability of the model.
To tackle this, we adopt the prediction probability $ \hat{y}_{i}^I $ and $ \hat{y}_{i}^T $ computed over $ k^\prime_{I_i} $ and $ k^\prime_{T_i} $ as soft attribute labels.
To further enhance the fine-grained alignment between image and text modalities, we use $ \hat{y}_{i}^I $ and $ \hat{y}_{i}^T $ as the soft label for $ p_{i}^T $ and $ p_{i}^I $, respectively, based on the assumption that matched image-text pairs should predict the same attributes.
The binary cross-entropy loss is organized as follows: \par
\begin{small} 
\vspace{-4mm}
\begin{equation}
\begin{split}
    \mathcal{L}_{A}^S = \sum_{j=1}^N \Big( & w_j\left( \hat{y}^T_{i, j}\log p_{i, j}^I + \left(1 - \hat{y}^T_{i, j}\right) \log \left(1 - p_{i, j}^I\right)\right) \\
                           + & w_j\left( \hat{y}^I_{i, j}\log p_{i, j}^T + \left(1 - \hat{y}^I_{i, j}\right) \log \left(1 - p_{i, j}^T\right)\right) \Big), \nonumber
  \label{equ:ac_loss}
\end{split}
\end{equation}
\end{small}
\begin{equation}
    \hat{y}_{i}^I = \sigma\left(W_k k^\prime_{I_i} + \bm{z}_k \right), \hat{y}_{i}^T = \sigma\left(W_k k^\prime_{T_i} + \bm{z}_k \right),
  \label{equ:ac_loss1}
\end{equation}
where $ W_k $ and $ \bm{z}_k $ is the momentum updating versions of $ W_q $ and $ \bm{z}_q $ .

Finally, the full pre-trained objective can be expressed as:
\vspace{-2mm}
\begin{small}
\begin{align}
   \mathcal{L} = \mathcal{L}_{I} + \beta\left(\mathcal{L}_{I2T} + \mathcal{L}_{T2I}\right) + \gamma\left(\left(1-\alpha\right)\mathcal{L}_{A}^H + \alpha \mathcal{L}_{A}^S\right),
  \label{equ:ac_loss}
\end{align}
\end{small}

\vspace{-4mm}
\noindent
where $ \alpha $, $ \beta $ and $ \gamma $ are hyper-parameters balancing the importance of different objectives.

\section{Experiments}

\subsection{Pretraing Implementation}
To ensure a fair comparison with previous approaches, we use a standard ResNet50~\cite{he2016deep} as our image encoder, while our text encoder is a 12-layer, 8-head transformer~\cite{vaswani2017attention} with a hidden size of 512, following CLIP. 
All experiments are conducted using the PyTorch framework~\cite{paszke2019pytorch} on 8 NVIDIA V100 GPUs. 
The model is trained from scratch with randomly initialized weights, using an AdamW optimizer~\cite{loshchilov2017decoupled} with a weight decay of 0.1. 
We train the model for 140 epochs, with a batch size of 2048. 
The learning rate is initialized at $1e^{-6}$ and gradually increased to $1e^{-3}$ after 7 epochs, before being decreased to $1e^{-6}$ using the linear decay strategy. 
The trainable hyper-parameter $\tau^\prime$ is initialized as 0.07.
For other parameters, unless otherwise stated, we set $K=65536$, $\tau=0.07$, $\alpha=0.2$, $\beta=0.5$, $\gamma=0.01$, and $M=1359$. 
We adopt two data augmentation strategies for different objectives. For image self-supervised contrastive loss, we use the same data augmentation strategy as \cite{fu2021unsupervised}. 
For image-text contrastive loss and multi-attribute classification loss, we use simple data augmentation techniques, including random horizontal flip and image normalization.

\subsection{Various Downstream Tasks}
We evaluate the capabilities of VAL-PAT on multiple pedestrian analysis tasks, including fully-supervised reID, unsupervised reID, occluded reID, person attribute recognition and text-based person search.

\begin{table*}[htb]
    \setlength{\tabcolsep}{3.3mm}
    \begin{subtable}[h]{0.5\textwidth}
        \centering
        \begin{tabular}{l|ccc}
        \shline
        pre-train & PCB~\cite{sun2018beyond} & BOT~\cite{luo2019bag} & MGN~\cite{wang2018learning} \\
        \hline
        INsup    & 74.6/86.7 & 76.4/86.4 & 79.4/89.0 \\ \hline
        LUP        & 78.2/87.3 & 79.2/88.5 & 82.1/91.0\\ \hline
        UP-ReID    & \textbf{79.0}/88.5 & \textbf{80.0}/\textbf{89.1} & 82.1/90.9 \\ \hline
        VAL-PAT   & 78.6/\textbf{89.5} & \textbf{80.0}/88.6 &  \textbf{83.0}/\textbf{91.5}\\ 
        \shline
        \end{tabular}
        \vspace{-1mm}
        \caption{DukeMTMC}
        \label{tab:super-duke}
    \end{subtable}
    \hfill
    \begin{subtable}[h]{0.5\textwidth}
    \centering
        \begin{tabular}{l|ccc}
        \shline
        pre-train & PCB~\cite{sun2018beyond} & BOT~\cite{luo2019bag} & MGN~\cite{wang2018learning} \\
        \hline
        INsup    & 80.9/92.7 & 86.7/94.8 & 85.3/94.3 \\ \hline
        LUP        & 80.7/92.9 & 86.5/94.6 & 85.8/94.2\\ \hline
        UP-ReID    & 81.7/93.2 & 88.0/95.3 & \textbf{89.7}/96.1 \\ \hline
        VAL-PAT   & \textbf{86.0/94.9} & \textbf{88.7}/\textbf{96.1} & 89.6/\textbf{96.3} \\ \shline
        \end{tabular}
        \vspace{-1mm}
        \caption{PersonX}
        \label{tab:super-personx}
    \end{subtable}
    \hfill
    \vspace{-2mm}
    \caption{Comparison of different pre-trained models on three representative fully-supervised reID methods.
The first number is mAP and the second is Rank-1.}
    \vspace{-1mm}
    \label{tab1}
\end{table*}

\begin{table*}[t]
    \setlength{\tabcolsep}{2.0mm}
    \begin{subtable}[h]{0.5\textwidth}
        \centering
        \begin{tabular}{l|ccc}
        \shline
        pre-train & SPCL~\cite{ge2020self} & RLCC~\cite{zhang2021refining} & C-Contrast~\cite{dai2022cluster} \\
        \hline
        INsup    & 65.3/81.2 & 69.2/83.2 & 72.6/84.9 \\ \hline
        LUP        & 67.1/81.6 & 69.3/82.6 & 72.7/84.7 \\ \hline
        UP-ReID    & 63.0/79.8 & 71.3/83.3 & 73.4/85.3  \\ \hline
        VAL-PAT   & \textbf{69.8/83.5} & \textbf{73.0}/\textbf{85.3} & \textbf{74.9/86.1} \\ \shline
        \end{tabular}
        \vspace{-1mm}
        \caption{DukeMTMC}
        \label{tab:unsup-duke}
    \end{subtable}
    \hfill
    \begin{subtable}[h]{0.5\textwidth}
    \centering
        \begin{tabular}{l|ccc}
        \shline
        pre-train & SPCL~\cite{ge2020self} & RLCC~\cite{zhang2021refining} & C-Contrast~\cite{dai2022cluster} \\
        \hline
       INsup    & 19.1/42.3 & 27.9/56.5 & 33.0/62.0 \\ \hline
        LUP        & 11.3/25.6 & 21.4/45.1 & 22.9/48.5\\ \hline
        UP-ReID    & 10.6/23.9 & 20.2/42.9 & 24.2/49.7 \\ \hline
        VAL-PAT   & \textbf{29.5}/\textbf{56.6} & \textbf{38.9}/\textbf{67.5} & \textbf{36.1/64.9} \\ \shline
        \end{tabular}
        \vspace{-1mm}
        \caption{MSMT17}
        \label{tab:unsup-msmt17}
    \end{subtable}
    \hfill
    \vspace{-2mm}
     \caption{Comparison of different pre-trained models on three representative unsupervised reID methods.
The first number is mAP and the second is Rank-1.}
    \label{tab2}
    \vspace{-4mm}
\end{table*}

\noindent
\textbf{Fully-supervised person reID} aims to associate images of a specified pedestrian from a large pool of images.
We conduct experiments on two reID datasets, DukeMTMC-reID~\cite{Ristani2016PerformanceMA} and PersonX~\cite{sun2019dissecting}\footnote{We only test on DukeMTMC-reID and PersonX, because CUHK-PEDES and ICFG-PEDES, which are used to train image captioning model, contain images from Market1501 and MSMT17, respectively.}.
DukeMTMC-reID contains 16,522 training images of 702 identities, 2,228 query images of the other 702 identities and 17,661 gallery images.
PersonX is a synthetic reID dataset that consists of 45,792 images of 1,266 identities from 6 cameras, with 9,840 images of 410 identities for training, 5,136 images of 856 identities for query and the other images for gallery.
We evaluate the performance using mean Average Precision (mAP) and cumulative matching characteristic~(CMC), as done in~\cite{fu2021unsupervised}. 
To validate the transferability of our VAL-PAT, we utilize three representative reID baselines: PCB~\cite{sun2018beyond}, BOT~\cite{luo2019bag}, and MGN~\cite{wang2018learning}.




\noindent
\textbf{Unsupervised Person reID} 
aims to learn discriminative pedestrian representations without annotations.
It can be categorized into unsupervised domain adaptation and unsupervised learning, depending on the availability of labeled source data during training. 
In this task, we focus on the setting where no labeled source data are available, and evaluate the performance on two widely used datasets: DukeMTMC-reID and MSMT17~\cite{wei2018person}\footnote{To test on MSMT17, we train image captioning model with only CUHK-PEDES and keep the other setting unchanged to avoid information leakage.}.
MSMT17 is the largest labeled reID dataset, containing 126,441 images of 4,101 identities.
We use the same metrics as in fully-supervised reID and select three state-of-the-art unsupervised reID methods, SpCL~\cite{ge2020self}, RLCC~\cite{zhang2021refining} and C-Contrast~\cite{dai2022cluster}, to evaluate the effectiveness of our pre-trained model.

\noindent
\textbf{Occluded Person reID} aims identify the same person with a full-body appearance in a disjoint camera, using an occluded query image.
We conduct experiments on Occluded-Duke~\cite{miao2019pose}, which is selected from DukeMTMC-reID by keeping occluded images and filtering out some overlapping images. 
It contains 15,618 training images, 17,661 gallery images, and 2,210 occluded query images.
We use the same metric as in fully-supervised reID and test the performance of different pre-trained models using HOReID~\cite{wang2020high}.


\noindent
\textbf{Person Attribute Recognition}
is a multi-label classification task, aiming at recognizing attributes presented in pedestrian images, such as the clothes style, clothes color, \emph{etc.}
To evaluate the performance of our VAL-PAT, we adopt the strong baseline RePAR~\cite{jia2021rethinking} (ResNet50 followed by an attribute classifier) and implement experiments on two popular datasets: RAP~\cite{li2016richly} and PA-100K~\cite{liu2017hydraplus}.
The RAP dataset contains 41,585 images, where each image is annotated with 72 attributes.
Following~\cite{li2016richly}, we split the PAR dataset into 33,268 training images and 8,317 testing images.
PA-100K contains 80,000 training images, 10,000 validation images and 10,000 test images, with each images annotated with 26 commonly used attributes.
Following commonly used setting, the mean accuracy~(mA), accuracy, precision, recall and F1 score are used for evaluation.

\noindent
\textbf{Text-based Person Search} 
utilizes textual descriptions as queries to retrieve relevant pedestrian images from a large gallery set.
We implement experiments on CUHK-PEDES dataset~\cite{li2017person}\footnote{To ensure a fair comparison with other methods in this task, we train image captioning model with only CUHK-PEDES and keep other setting unchanged.}
, which is comprised of 40,206 pedestrian images of 13,003 identities with 80,412 textual descriptions. 
The total dataset is split into 34,054 images of 11,003 identities for training, 3,078 images of 1,000 identities for validation and 3,074 images of 1,000 identities for testing.
We adopt the top-$ K $ accuracy metric~($ K $=1, 5, 10) for evaluation, following the standard setting. 
Our baseline method for this task is NAFS~\cite{gao2021contextual}, and we reimplement it with a better data augmentation strategy~\cite{shu2023see}, resulting in improved performance compared to the original paper.

\subsection{Comparing with Pre-trained Methods}
In this section, we compare VAL-PAT with other pre-trained methods based on ResNet50 backbone, including INsup~(a supervised model pre-trained on ImageNet), 
LUP~(pre-training MoCo v2 on LUPerson) and UP-reID~(a pre-trained model elaborately designed for reID task).
To ensure a fair comparison, we search for optimal hyper-parameters for each pre-trained model to present the best performance during fine-tuning on downstream tasks.

\noindent
\textbf{Fully-supervised person reID.}
Table~\ref{tab1} presents the improvements using the three representative fully-supervised reID methods on the two popular person reID datasets.
In contrast to our baseline LUP, the MGN with VAL-PAT obtain \textbf{0.9}\% and \textbf{3.8}\% improvements in terms of mAP and \textbf{0.5}\% and \textbf{2.1}\% improvements in terms of Rank-1 on DukeMTMC and PersonX, respectively.
Although UP-ReID is a sophisticated pre-trained model for reID tasks, our VAL-PAT still achieves competitive performance on DukeMTMC and PersonX.
Notably, the UP-ReID framework can be used to replace the vanilla image self-supervised branch in VAL-PAT, which has the potential to further improve the performance on downstream tasks.

\noindent
\textbf{Unsupervised person reID.}
Table~\ref{tab2} demonstrates that our pre-trained model is more effective in unsupervised reID settings than in fully-supervised reID.
Generally, our VAL-PAT consistently outperforms other pre-trained models by a considerable margin with different baselines on DukeMTMC and MSMT17.
Specifically, the C-Contrast with our VAL-PAT surpasses the one with UP-reID by \textbf{1.5}\%, \textbf{11.9}\% in terms of mAP and \textbf{0.8}\%, \textbf{15.2}\% in terms of Rank-1 on DukeMTMC and MSMT17, respectively.
Moreover, it is observed that the performance of LUP and UP-ReID is far below INsup on MSMT17, but our VAL-PAT significantly exceeds INsup.
This indicates that our VAL-PAT trained with multimodal supervision can learn more discriminative representations.

\noindent
\textbf{Occluded person reID.}
As shown in Table~\ref{tab:occ-reid}, 
our VAL-PAT can also benefit occluded reID tasks.
Specifically, compared to initializing with LUP, HOReID with VAL-PAT achieves \textbf{2.4}\% and \textbf{2.6}\% improvements in terms of mAP and Rank-1.
Additionally, VAL-PAT also outperforms UP-ReID by \textbf{1.6}\% on mAP and \textbf{2.4}\% on Rank-1, showing the effectiveness of our pre-training framework.
\begin{table}[h!]
\small
\vspace{-3mm}
\centering
\setlength{\tabcolsep}{5.4mm}
    \begin{tabular}{l|cc}
    \shline
    \centering
    \multirow{2}{*}{pre-train} & \multicolumn{2}{c}{Occluded-Duke}  \\
    \cline{2-3} & mAP & Rank-1  \\
    \hline
    \centering
    INsup     & 43.8 &  55.1 \\
    \centering
    \centering
    LUP  & 51.5 & 60.3  \\
    \centering
    UP-ReID  & 52.3 & 60.5  \\
    \centering
    VAL-PAT & \textbf{53.9} & \textbf{62.9} \\
    \shline
    \end{tabular}
    \vspace{-2mm}
    \caption{Comparison of different pre-trained models on occluded reID.
    HOReID~\cite{wang2020high} is adopted as baseline method.}
    \label{tab:occ-reid}
    \vspace{-2mm}
\end{table}

\noindent
\textbf{Pedestrian Attribute Recognition.}
As illustrated in Table~\ref{tab:par}, our VAL-PAT consistently outperforms other pre-trained models in terms of mA and F1 on both RAP and PA-100K datasets.
Compared with our VAL-PAT, the LUP and UP-reID only achieve slight performance improvements over INsup on both datasets, indicating that these pre-trained models specifically designed for reID tasks lack the ability to generalize well to other pedestrian analysis tasks.
\begin{table}[h!]
\vspace{-1mm}
\small
    \centering
    \setlength{\tabcolsep}{3.5mm}
    \begin{tabular}{l|cc|cc}
    \shline
    \centering
    \multirow{2}{*}{pre-train} & \multicolumn{2}{c|}{RAP} & \multicolumn{2}{c}{PA-100K} \\
    \cline{2-5} & mA & F1 & mA & F1 \\
    \hline
    \centering
    INsup    & 80.2 & 80.5 & 80.9 & 87.4 \\
    \centering
    \centering
    LUP  & 80.6 & 80.7 & 81.1 & 88.1 \\
    \centering
    UP-ReID  & 80.5 & 80.5 & 81.1 & 88.1 \\
    \centering
    VAL-PAT & \textbf{80.8} & \textbf{81.0} & \textbf{82.3} & \textbf{88.5} \\
    \shline
    \end{tabular}
    \vspace{-1mm}
    \caption{Comparison of different pre-trained models on pedestrian attribute recognition.
    RePAR~\cite{jia2021rethinking} is adopted as baseline method.}
    \vspace{-2mm}
    \label{tab:par}
\end{table}

\noindent
\textbf{Text-based person search.}
We compare the VAL-PAT with other pre-trained models on CUHK-PEDES and present the results in Table~\ref{tab:tps}.
Since text-based person search is a multimodal retrieval task, the text encoder is required to extract text representations.
For INsup, LUP and UP-ReID, we adopt BERT~\cite{devlin2018bert} as their text encoder.
In contrast, our VAL-PAT directly utilizes the text encoder from pre-training phase.
We can observe that our VAL-PAT outperforms LUP and UP-ReID by \textbf{2.7}\% and \textbf{2.3}\% on top-1 accuracy, respectively.
The superiority of our pre-trained model can be attributed to the fact that it has established a strong relationship between the image and text modalities through the use of image-text contrastive loss and multi-attribute classification loss during pre-training, making it easily adaptable to this task.
\begin{table}[h!]
\vspace{-1mm}
\small
    \centering
    \setlength{\tabcolsep}{4.5mm}
        \begin{tabular}{l|ccc}
        \shline
        \centering
        \multirow{2}{*}{pre-train} & \multicolumn{3}{c}{CUHK-PEDES}  \\
        \cline{2-4} & top-1 & top-1 & top-10  \\
        \hline
        \centering
        INsup   & 61.2 & 80.6 & 87.2\\
        \centering
        \centering
        LUP  &  62.0 & 81.2 &  87.7\\
        \centering
        UP-ReID  & 62.4 &  81.4 & 88.0\\
        \centering
        VAL-PAT & \textbf{64.7} & \textbf{83.3} & \textbf{88.9}\\
        \shline
        \end{tabular}
        \vspace{-1mm}
        \caption{Comparison of different pre-trained models on text-based person search.
        NAFS~\cite{gao2021contextual} is used as our baseline method.}
        \vspace{-5mm}
        \label{tab:tps}
\end{table}

\begin{table}[t]
\small
\centering
\setlength{\tabcolsep}{1.5mm}
\begin{tabular}{l|cc|cc}
    \shline
    \multirow{2}{*}{Methods} & \multicolumn{2}{c|}{DukeMTMC} & \multicolumn{2}{c}{Occluded-Duke} \\
    \cline{2-5} & mAP & Rank-1 & mAP & Rank-1 \\ 
    \hline
    PCB~\cite{sun2018beyond} (ECCV'18) & 69.2 & 83.3 & 33.7 & 42.6 \\
    MGN~\cite{wang2018learning} (MM'18) & 78.4 & 88.7 & - & - \\
    MGN*        & \underline{79.4} & \underline{89.0} & - & - \\
    BOT~\cite{luo2019bag} (CVPRW'19) & 76.4 & 86.4 & - & - \\
    BDB~\cite{dai2019batch} (ICCV'19) & 76.0 & 89.0 & - & - \\
    PGFA~\cite{miao2019pose} (ICCV'19) & 65.5 & 82.6 & 37.3 & 51.4 \\
    SAN~\cite{jin2020semantics} (AAAI'20) & 75.5 & 87.9 & - & - \\
    ISP~\cite{zhu2020identity} (ECCV'20) & 80.0 & 89.6 & 52.3 & 62.8 \\
    GASM~\cite{he2020guided} (ECCV'20) & 74.4 & 88.3 & - & - \\
    PVPM~\cite{gao2020pose} (CVPR'20) & - & - & 37.7 & 47.0 \\
    HOReID~\cite{wang2020high} (CVPR'20) & 75.6 & 86.9 &  \underline{43.8} & \underline{55.1} \\
    AMD~\cite{chen2021explainable} (ICCV'21) & 75.5 & 88.2 & - & - \\
    OAMN~\cite{chen2021occlude} (ICCV'21) & 72.6 & 86.3 & 46.1 & 62.6 \\
    PAT~\cite{li2021diverse} (CVPR'21) & 78.2 & 88.8 & 53.6 & \textbf{64.5} \\
    \hline
    VAL-PAT~(Ours) & \textbf{83.0} & \textbf{91.5} & \textbf{53.9} & 62.9 \\
    \shline
\end{tabular}\\
\vspace{-1mm}
\caption{Performance (\%) comparisons with state-of-the-art fully-supervised reID approaches on DukeMTMC-reID and occluded reID approaches on Occluded-Duke.
* refers to the re-implementation based on official released code.
The best performance is marked as bold and our baseline is highlighted with underline.}
\vspace{-2mm}
\label{tab4}
\end{table}

\begin{table}
\small
\centering
\setlength{\tabcolsep}{1.4mm}
\begin{tabular}{l|cc|cc}
    \shline
    \multirow{2}{*}{Methods} & \multicolumn{2}{c|}{DukeMTMC} & \multicolumn{2}{c}{MSMT17} \\
    \cline{2-5} & mAP & Rank-1 & mAP & Rank-1 \\ 
    \hline
    MMCL~\cite{wang2020unsupervised} (CVPR'20) & 40.2 & 65.2 & 11.2 & 35.4 \\
    HCT~\cite{zeng2020hierarchical} (CVPR'20) & 50.7 & 69.6 & - & - \\
    MMT~\cite{dubourvieux2021unsupervised} (ICLR'20) & 60.3 & 75.6 & - & - \\
    SpCL~\cite{ge2020self} (NeurIPS'20) & 65.3 & 81.2 & 19.1 & 42.3 \\
    GCL~\cite{chen2021joint} (CVPR'21) & 62.8 & 82.9 & 21.3 & 45.7 \\
    RLCC~\cite{zhang2021refining} (CVPR'21) & 69.2 & 83.2 & \underline{27.9} & \underline{56.5} \\
    C-Contrast~\cite{dai2022cluster} (ACCV'22) & \underline{72.6} & \underline{84.9} & 33.0 & 62.0 \\
    PPLR~\cite{cho2022part} (CVPR'22) & - & - & 31.4 & 61.1 \\
    ISE~\cite{zhang2022implicit} (CVPR'22) & - & - & 37.0 & \textbf{67.6} \\
    \hline
    VAL-PAT~(Ours) & \textbf{74.9} & \textbf{86.1} & \textbf{38.9} & 67.5 \\
    \shline
\end{tabular}\\
\vspace{-1mm}
\caption{Performance (\%) comparisons with state-of-the-art unsupervised reID approaches on DukeMTMC and MSMT17. }
\label{tab5}
\vspace{-6mm}
\end{table}

\subsection{Comparing with State-of-the-Art Methods}
In this section, we compare our results with the state-of-the-art methods in aforementioned downstream pedestrian-related tasks 
and report these results in Table~\ref{tab4}, Table~\ref{tab5}, Table~\ref{tab6} and Table~\ref{tab7}.
As shown in Table~\ref{tab4}, using our pre-trained weights, the MGN achieves the best performance in terms of mAP and Rank-1 on DukeMTMC-reID for fully-supervised reID tasks.
Additionally, the HOReID with our VAL-PAT surpasses the state-of-the-art PAT by \textbf{0.3}\% in terms of mAP on Occluded-Duke.
In Table~\ref{tab5}, for the unsupervised reID, we achieve a new state-of-the-art performance on DukeMTMC and MSMT17 by simply applying our pre-training ResNet50 model on C-Contrant and RLCC, respectively.
In Table~\ref{tab6}, the simplest baseline method RePAR~(ResNet50 followed by an attribute classifier) with our VAL-PAT has exceeded all sophisticated methods except for Recall metric on PA-100K. 

Moreover, as presented in Table~\ref{tab7}, the NAFS with our VAL-PAT outperforms the state-of-the-art LGUR by \textbf{0.5}\% in terms of top-1 accuracy on CUHK-PEDES, thanks to our multimodal learning framework.
In summary, the consistent improvements on these tasks prove that our pre-trained model can learn transferable representations that boost most of pedestrian-related tasks. 

\begin{table*}[h!]
\small
\centering
\setlength{\tabcolsep}{2.67mm}
\begin{tabular}{l|cc|cc|ccccc|cc}
    \shline
    \multirow{2}{*}{pre-train} & \multicolumn{2}{c|}{DukeMTMC} & \multicolumn{2}{c|}{MSMT17} & \multicolumn{5}{c|}{PA-100K} & \multicolumn{2}{c}{CUHK-PEDES} \\
    \cline{2-12} & mAP & Rank-1 & mAP & Rank-1 & mA & Accu & Prec & Recall & F1 & top-1 & top-10\\ 
    \hline
    INsup  & 76.4 & 86.4 & 27.9 & 56.5 & 80.9 & 79.2 & 87.7 & 87.2 & 87.4 & 61.2 & 87.2  \\
      SSL  & 77.9 & 87.3 & 18.7 & 38.4 & 81.2 & 80.4 & 88.2 & 88.3 & 88.3 & 61.5 & 87.2  \\
      ITC  & 78.6 & 87.8 & 35.5 & 64.1 & 81.6 & 79.6 & 87.8 & 87.6 & 87.7 & 62.9 & 88.2 \\
      SSL+ITC        & 79.6 & \textbf{88.9} & 37.7 & 66.3 & 82.1 & 80.7 & \textbf{88.5} & 88.3 & 88.4 & 63.5 & 88.6 \\
      SSL+ITC+MAC \footnotesize{w/o soft}  & 78.1 & 88.1 & 34.0 & 62.2 & 81.9 & 80.5 & 88.3 & 88.1 & 88.2 & 62.5 & 87.8 \\
      SSL+ITC+MAC  & \textbf{80.0} & 88.6 & \textbf{38.9} & \textbf{67.5} & \textbf{82.3} & \textbf{80.9} & \textbf{88.5} & \textbf{88.5} & \textbf{88.5} & \textbf{64.7} & \textbf{88.9} \\
    \shline
\end{tabular}\\
\vspace{-2mm}
\caption{Ablating the different components of VAL-PAT pre-training w.r.t downstream task performances. }
\label{tab8}
\vspace{-3mm}
\end{table*}

\begin{table}[t]
\small
\centering
    \setlength{\tabcolsep}{1.1mm}{
    \begin{tabular}{l|ccccc}
    \shline
    Methods & mA & Accu & Pre & Recall & F1\\
    \hline
    VeSPA~\cite{sarfraz2017deep} (BMVC'17) & 76.3 & 73.0 & 85.0 & 81.5 & 83.2 \\
    LGNet~\cite{liu2018localization} (BVMC'18) & 77.0 & 75.6 & 87.0 & 83.2 & 85.0 \\
    PGDM~\cite{li2018pose} (ICME'18) & 75.0 & 73.0 & 84.4 & 82.2 & 83.3 \\
    MsVAA~\cite{sarafianos2018deep} (ECCV'18) & 80.1 & 77.0 & 86.2 & 85.6 & 85.5\\
    VAC~\cite{guo2019visual} (CVPR'19) & 79.0 & 79.0 & 88.4 & 86.1 & 86.8 \\
    ALM~\cite{tang2019improving} (ICCV'19) & 79.3 & 78.6 & 87.3 & 86.7 & 86.6 \\
    RePAR~\cite{jia2021rethinking} (ArXiv'21) & 80.2 & 79.2 & 87.8 & 87.0 & 87.4 \\
    RePAR* & \underline{80.9} & \underline{79.2} & \underline{87.7} & \underline{87.2} & \underline{87.4} \\
    SSC~\cite{jia2021spatial} (ICCV'21) & 81.9 & 78.9 & 86.0 & \textbf{89.1} & 86.9\\
    Label2Label~\cite{li2022label2label} (ECCV'22) & 82.2 & 79.2 & 86.4 & 88.6 & 87.1\\
    \hline
    VAL-PAT~(Ours) & \textbf{82.3} & \textbf{80.9} & \textbf{88.5} & 88.5 & \textbf{88.5}\\
    \shline
\end{tabular}}\\
\caption{Performance (\%) comparisons with state-of-the-art pedestrian attribute recognition approaches on PA-100K. }
\label{tab6}
\vspace{-4mm}
\end{table}

\subsection{Ablation Study}
We analyze the effectiveness of different components in our VAL-PAT through ablation studies on four downstream tasks: fully-supervised reID, unsupervised reID, pedestrian attribute recognition and text-based person search.
For these tasks, we adopt BOT, RLCC, RePAR and NAFS as the baseline methods, respectively. 
The complete results are listed in Table~\ref{tab8}, where the SSL refers to image self-supervised contrastive loss, the ITC denotes image-text contrastive loss and the MAC stands for multi-attribute classification loss.
We first validate the effectiveness of pre-training objectives SSL and ITC by testing their standalone performance.
As shown in Tabel~\ref{tab8}, with the exception of SSL in unsupervised setting, both per-training objectives contribute to downstream tasks compared to the INsup.

In contrast to using SSL alone, we observe significant improvements on all downstream tasks after adding ITC~(SSL+ITC), demonstrating that language supervision can effectively guide visual representation learning by providing detailed and accurate information about pedestrians' appearance. 
We further investigate the impact of adding MAC to SSL+ITC, but only using mined noisy labels as attribute supervision~(SSL+ITC+MAC w/o soft).
As expected, the performances on downstream tasks are even worse than SSL+ITC, supporting our hypothesis that enforcing the model to fit these noisy labels impairs the generalization ability of the representations.
However, with the incorporation of soft labels as a complement to the mined noisy attributes (SSL+ITC+MAC), we observe notable improvements in the performance across various downstream tasks, particularly in pedestrian attribute recognition and text-based person search, as compared to SSL+ITC.
This proves that MAC not only facilitates the learning of fine-grained visual representations but also helps bridge the modality gap between image and text by encouraging the model to recognize common attributes that are presented in matched image-text pairs.
In conclusion, the best results in SSL+ITC+MAC illustrate these three objectives, SSL, ITC and MAC, are highly synergistic and guide the model to learn more discriminative visual representations.

\begin{table}[t]
\small
\centering
    \setlength{\tabcolsep}{3.4mm}{
    \begin{tabular}{l|ccc}
    \shline
    Methods & top-1 & top-5 & top-10 \\
    \hline
    TIMAM~\cite{sarafianos2019adversarial} (ICCV'19) & 54.5 & 77.6 & 84.8  \\
    ViTAA~\cite{wang2020vitaa} (ECCV'20) & 56.0 & 75.8 & 83.5  \\
    CMAAM~\cite{aggarwal2020text} (WACV'20) & 56.7 & 77.2 & 84.9  \\
    HGAN~\cite{zheng2020hierarchical} (MM'20) & 59.0 & 79.5 & 86.6\\
    NAFS~\cite{gao2021contextual} (ArXiv'21) & 59.9 & 79.9 & 86.7  \\
    NAFS* & \underline{61.2} & \underline{80.6} & \underline{87.2}  \\
    DSSL~\cite{zhu2021dssl} (MM'21) & 60.0 & 80.4 & 87.6  \\
    SSAN~\cite{ding2021semantically} (ArXiv'21) & 61.4 & 80.2 & 86.7\\
    SRCF~\cite{suo2022simple} (ECCV'22) & 64.0 & 83.0 & 88.8\\
    LGUR~\cite{shao2022learning} (MM'22) & 64.2 & 81.9 & 87.9\\
    \hline
    VAL-PAT~(Ours) & \textbf{64.7} & \textbf{83.3} & \textbf{88.9} \\
    \shline
\end{tabular}}\\
\caption{Performance (\%) comparisons with state-of-the-art text-based person search approaches on CUHK-PEDES. }
\label{tab7}
\vspace{-4mm}
\end{table}

\vspace{-2mm}
\section{Conclusion}
In this paper, we present a novel multimodal learning framework VAL-PAT that learns transferable pedestrian representation from multimodal information supervision.
We first introduce a multimodal pedestrian dataset, LUPerson-TA, with text and attribute annotations, and then train the framework on this dataset using image self-supervised contrastive learning, image-text contrastive learning and multi-attribute classification.
Extensive experiments show that the representation learned by VAL-PAT can transfer well to downstream pedestrian analysis tasks, including person reID, pedestrian attribute recognition, and text-based person search.
Compared to previous pre-trained models, our VAL-PAT achieves superior transfer performance.
Moreover, our proposed framework even outperforms previous state-of-the-art methods on these tasks.


{\small
\bibliographystyle{ieee_fullname}
\bibliography{egbib}
}

\newpage
\appendix
\noindent\textbf{\Large Appendix}

\begin{figure*}[tp]
	\centering
	\includegraphics[width=\linewidth]{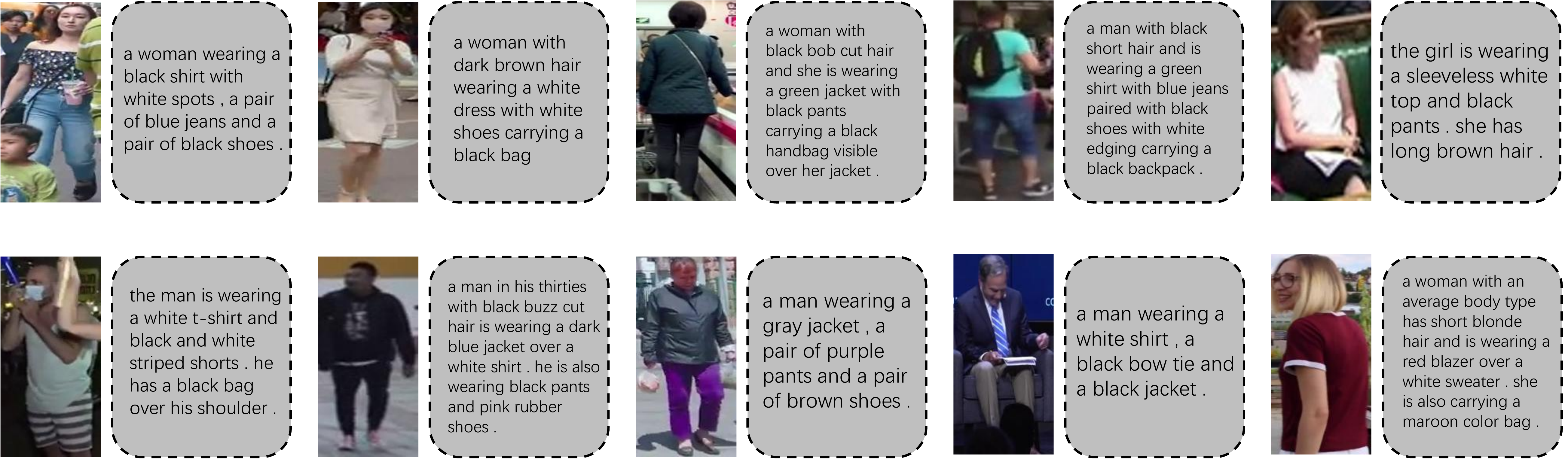}
	\caption{
		Additional image-text pairs in LUPerson-TA dataset.
		The textual descriptions in the first row are accurate, whereas the descriptions in the second row are noisy and imprecise.
	}
	\label{app:fig1} 
\end{figure*}

\begin{table*}[tp]
	\small
	\centering
	\setlength{\tabcolsep}{3.3mm}
	\begin{tabular}{l|cc|ccccc|ccc}
		\shline
		\multirow{2}{*}{pre-train} & \multicolumn{2}{c|}{Occluded-Duke} &  \multicolumn{5}{c|}{RAP} & \multicolumn{3}{c}{CUHK-PEDES} \\
		\cline{2-11} & mAP & Rank-1 & mA & Accu & Prec & Recall & F1 & top-1 & top-5 & top-10\\ 
		\hline
		INsup  & 43.8 & 55.1 & 80.2 & 68.6 & 79.9 & 81.1 & 80.5 & 56.9 &  77.4 & 84.2   \\
		SSL  & 51.8 & 60.8 & 79.5 & 68.4 & 79.7 & 80.1 & 80.3 & 58.4 & 78.6 & 85.0   \\
		ITC  & 49.6 & 59.8 & 80.1 & 68.7 & 80.1 & 81.1 & 80.6 & 60.3 & 79.5  & 85.9  \\
		SSL+ITC        & 53.7 & \textbf{63.2} & 80.6 & 69.2 & 80.3 & 81.5 & 80.9 & 61.3 & 80.0 & 87.0 \\
		SSL+ITC+MAC \footnotesize{w/o soft}  & 52.6 & 62.4 & 80.4 & \textbf{69.3} & \textbf{80.5} & 81.4 & \textbf{81.0} & 61.0 & 79.8 & 86.3  \\
		SSL+ITC+MAC  & \textbf{53.9} & 62.9 & \textbf{80.8} & \textbf{69.3}  & 80.4 & \textbf{81.6} & \textbf{81.0} & \textbf{62.4} & \textbf{80.9}  & \textbf{88.3}  \\
		\shline
	\end{tabular}\\
	\vspace{-2mm}
	\caption{Ablating the different components of VAL-PAT pre-training w.r.t downstream task performances. 
		For INsup and SSL, we adopt Bert~\cite{devlin2018bert} as their text encoder.
		For other setting, we direct use text encoder from the pre-training phase.}
	\label{app:tab1}
	\vspace{2mm}
\end{table*}

\section{More Image-text pairs in LUPerson-TA}
Figure~\ref{app:fig1} demonstrates additional image-text pairs from the LUPerson-TA dataset. 
The textual descriptions of the image-text pairs in the first row accurately describe the characteristics of the corresponding pedestrians, including fine-grained details such as ``white dots" in row 1, column 1, ``a black handbag" in row 1, column 3, and ``long brown hair" in row 1, column 5, which are advantageous for learning more discriminative visual representations. 
However, due to the use of a simple image captioning method~\cite{xu2015show} and the limited training samples (a total of 102,782 image-text pairs in CUHK-PEDES~\cite{li2017person} and ICFG-PEDES~\cite{ding2021semantically}), some textual descriptions generated by the image captioning model contain noise, such as incorrect recognition of the color of the pedestrian's shirt in row 2, column 2, and omission of the descriptions of the pedestrian's pants in row 2, column 4. 
These imprecise textual descriptions not only hinder the convergence of image-text contrastive loss, but also introduce noise into the mined attribute labels. 
In this paper, we propose optimization objective MAC to alleviate this issue. 
Alternatively, as a future research direction, we may explore more sophisticated image captioning methods to generate more accurate textual descriptions.


\section{More Ablation Study}
We present more ablation studies on occluded reID, pedestrian attribute recognition and text-based person search, and report these results in Table~\ref{app:tab1}.
For these tasks, we respectively select HOReID~\cite{wang2020high}, RePAR~\cite{jia2021rethinking} and CMPM/C~\cite{zhang2018deep}~(we reimplement CMPM/C with better data augemtation strategies) to evaluate performance of different variants of VAL-PAT.
As can be seen, the combination of SSL+ITC+MAC~(our VAL-PAT) achieves the best performances on most of metrics across these tasks.

\section{Visualizing the Pre-trained Image Encoder}
\begin{figure}[H]
	\centering
	\includegraphics[width=\linewidth]{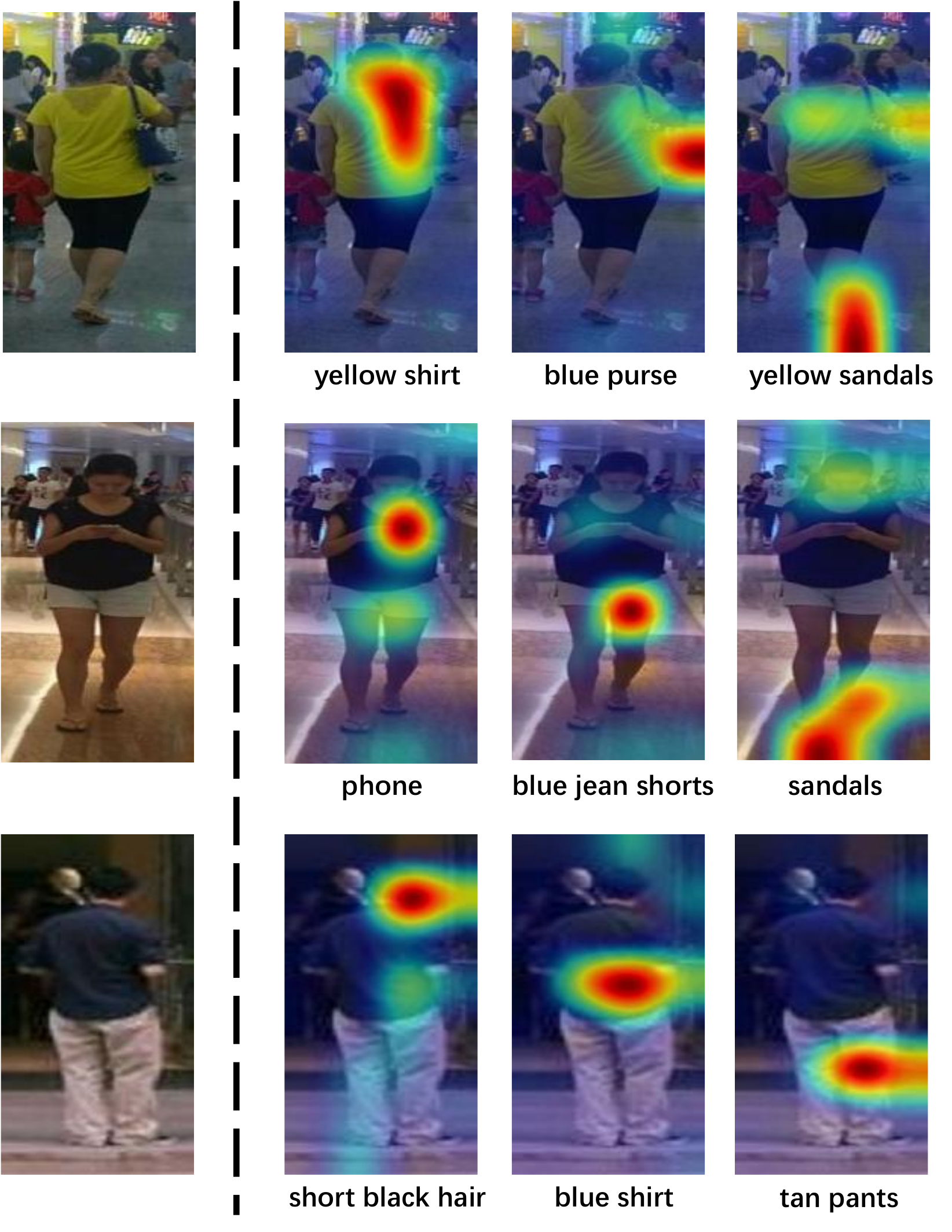}
	\caption{
		Grad-CAM visualizations of the image encoder $ E^I_q $, given different text queries.
	}
	\vspace{-2mm}
	\label{app:fig2} 
	\vspace{-2mm}
\end{figure}

In Figure~\ref{app:fig2}, we show the Grad-CAM~\cite{selvaraju2017grad} visualizations for the image encoder $ E_q^I $ in our VAL-PAT, with different text queries fed into the text encoder $ E_q^T $.
Remarkably, we observe that our image encoder successfully localizes the relevant regions for various query texts, even for fine-grained queries such as ``phone" and ``short black hair".
This indicates our pre-trained framework effectively captures a wide range of visual concepts and establishes strong associations between these concepts and their corresponding texts. 

\end{document}